# Celcomen: spatial causal disentanglement for single-cell and tissue perturbation modeling

Stathis Megas[1,2,3*†], Daniel G. Chen[1,2,4†], Krzysztof Polanski[1,2], Moshe Eliasof[4], Carola Bibiane Schönlieb[4], Sarah A. Teichmann[1,2,3,5*]

[1] Department of Cellular Genetics, Wellcome Sanger Institute, Hinxton, United Kingdom
[2] Cambridge Stem Cell Institute, Department of Medicine, University of Cambridge, Cambridge, United Kingdom
[3] Cambridge Center for AI in Medicine, Department of Applied Mathematics and Theoretical Physics, University of Cambridge, Cambridge, United Kingdom
[4] Department of Applied Mathematics and Theoretical Physics, University of Cambridge, Cambridge, United Kingdom
[5] Canadian Institute for Advanced Research, Toronto, Canada
† Equal contribution
* Corresponding authors

## Abstract
Celcomen leverages a mathematical causality framework to disentangle intra- and inter-cellular gene regulation programs in spatial transcriptomics and single-cell data through a generative graph neural network. It can learn gene-gene interactions, as well as generate post-perturbation counterfactual spatial transcriptomics, thereby offering access to experimentally inaccessible samples. We validated its disentanglement, identifiability, and counterfactual prediction capabilities through simulations and in clinically relevant human glioblastoma, human fetal spleen, and mouse lung cancer samples. Celcomen provides the means to model disease and therapy induced changes allowing for new insights into single-cell spatially resolved tissue responses relevant to human health.

## Main
A cell's gene expression profile simultaneously encodes information about its intrinsic characteristics and extrinsic tissue microenvironment. Recent technologies now allow for large-scale profiling of transcriptomes at single-cell resolution with spatial context[1–4]. With these technological advances, computational methods that can disentangle intrinsic versus extrinsic inter-cellular regulation of gene expression are needed. These disentangled representations are necessary to fully reconstruct the complex interplay of intra- and inter- cellular interactions in human tissues during homeostasis and post-disease or therapy induced perturbation[5,6].

Several previous works relied on prior knowledge of protein-protein interactions (PPI) or gene regulatory networks (GRN) to distinguish intrinsic and extrinsic circuits; this reliance often excludes key cell-cell interaction partners that are unreported[7,8]. Recent deep learning models advance on this limitation by simultaneously modeling intrinsic and extrinsic features; however these models lack interpretable insight due to their black box nature[9]. Further, most current models lack mathematical (identifiability) guarantees, leading to their hyper-sensitivity to input data variability; exceedingly few accept both spatial and single-cell input data[10,11]; and many cannot perform *in silico* perturbation experiments critical to understanding tissue behavior during disease[12–14]. While these works have introduced marked computational leaps in spatial transcriptomics data interpretation, they often cannot perform causal inferences due to their lack of identifiability which mathematically prevents many current models from deriving comprehensive mechanistic insights into cell and tissue biology.

Celcomen overcomes some of these limitations by leveraging a causally identifiable framework into a generative graph neural network for learning disentangled representations of intra- and inter- cellular gene regulation in spatial transcriptomics data (**Fig. 1a-b**). The inference module of Celcomen, hereby called CCE, finds disentangled representations of gene interactions at the cell, 1st neighbor, 2nd neighbor, etc. levels. These representations can then be used by the generative module of Celcomen (Simcomen), hereby called SCE, to produce single-cell spatially resolved predictions of tissue behavior post perturbation and to derive realistic slides of spatial transcriptomics data from noise, ideologically similar to deep learning diffusion models. We validated the robustness and insightfulness of CCE and SCE across a plethora of simulations, input types, and in human tissues. In summary, we demonstrate Celcomen as a mathematically grounded spatial and single cell transcriptomics analysis tool that introduces the capability to perform high-resolution spatially resolved perturbation predictions that are critical for clinically relevant disease modeling and tissue engineering efforts.

**Celcomen's mathematical identifiability guarantees are reproduced in simulations**
Causal inference frameworks seek to uncover the mechanisms that generate the observed data by leveraging the mathematical principle of identifiability[15–18]. This principle holds true when there exists a single unique model and parameter combination that fits the data, and thus we are assured that our observations can be explained by the given model. However, despite the useful properties that are enjoyed by these identifiable models (e.g. robustness, generalizability, and self-consistency), most current deep learning models violate this principle[16,19]. We overcome this limitation through mathematical proofs of Celcomen's identifiability (**see Supplemental Notes**).

To confirm that Celcomen's identifiability guarantees exist in practice, we subjected Celcomen to a multitude of self-consistency simulations (**see Methods**). First, we randomly generated a ground truth set of gene-gene interactions. Next, we utilized Celcomen's generative module, SCE, to generate spatial transcriptomics data representative of these gene-gene interactions. Then, we fed the generated data into Celcomen's inference module, CCE, in an attempt to retrieve the originally encoded gene-gene interaction forces. In agreement with its identifiability guarantee, Celcomen consistently demonstrated strong alignment between its inferred gene-gene interactions from its simulated data and the ground truth (**Fig. 1c**). This suggests that Celcomen possesses strong self-consistency, and thus identifiability, as it is able to move between encoded gene-gene interactions to simulated spatial transcriptomics and then back to inferred gene-gene interactions with minimal, if not no, loss of information.

To confirm Celcomen's identifiability guarantees on *ex vivo* human data, we applied our model to multiple spatial transcriptomics slides of human fetal spleen[20]. For each slide, we trained a sample specific model and a model trained on the remaining samples. We then correlated the gene-gene interaction matrices of these two models. In line with its claimed identifiability, we observed strong positive correlation between these two gene-gene interaction matrices even though they shared no training samples (**Extended Data Figure 1**). Thus, through this computational experiment, we demonstrate that Celcomen's identifiability, and thus stability and robustness, extends beyond theory and simulations and can also be observed, and its rewards reaped, on human tissue sections.

**Celcomen recapitulates expected immune programs post interferon perturbation**
Having confirmed Celcomen's robustness as a model through simulation-based testing of its identifiability guarantees (**Fig. 1**), we then sought to test its value and validity in disentangling intra- versus inter- cellular gene regulation programs and in performing spatially resolved perturbation modeling. To test these claims, we applied Celcomen in a real human clinical setting by analyzing a single-cell resolution spatial transcriptomics dataset of human glioblastoma (brain cancer) (**Fig. 2a**). Consistent with its core theory, we found that Celcomen was able to successfully disentangle intrinsic versus extrinsic sources of transcriptomic variation through its assignment of gene-gene interactions involving secreted genes as inter-cellular, and those solely involving cytoplasmic genes as intra-cellular (**Fig. 2b**). It is important to note that the knowledge of which genes are secreted and which are cytoplasmic is not encoded into the model as prior information, but rather is learned by the model in an unsupervised manner.

We leveraged Celcomen's perturbation abilities to model interferon signaling in the context of a neurological tumor, where we investigate the scenario of interferon knockout. We chose to model interferon signaling due to its critical role in cancer in inducing antigen presentation, inflammation, and immune activation[21–23]. First, we quantified the expression of our sample's interferon associated gene program by averaging differentially upregulated genes in interferon (*IFITM3*) high versus low cells. Next, we knocked out interferon expression in a randomly chosen interferon high cell (**Fig. 2c**). Utilizing this interferon score, we not only confirmed our *in silico* knockout of interferon in the perturbed cell, by observing its marked loss of interferon associated genes, but we also observed loss of interferon signaling in neighbors of the perturbed cell (**Fig. 2d**). This behavior is highly consistent with known interferon biology as interferon signaling physically propagates from cell-to-cell within human tissues; thus recapitulating this intercellular signaling phenomenon supports the validity of Celcomen's perturbation modeling[24–26].

To further confirm the validity of our interferon knockout modeling, we performed pathway enrichment on genes that were differentially changed in perturbed (and perturbed neighboring) compared to unperturbed cells (**see Methods**). Indeed, we find that post interferon knockout, perturbed cells and their neighbors significantly downregulated characteristic interferon response programs compared to unperturbed cells (**Fig. 2e**). For example, we observed the perturbed cells to have decreased T cell effector and activation gene programs, as well as greater loss of infection-related gene sets and marked increases in regulatory programs. The consistency of our model with multiple aspects of known interferon biology strongly affirms Celcomen's ability to model perturbations with spatial resolution. Thus, through an in-depth case study of Celcomen on *ex vivo* human tissue sections, we provide significant validation to its value in disentangling intra- versus

inter- cellular gene regulation programs, and in performing high-resolution spatially contextualized perturbation modeling with accuracy.

**Celcomen spatial perturbation predictions are validated *in vivo***
Having validated Celcomen with *ex vivo* human tissue, we sought to *in vivo* validate its spatial perturbation modeling ability using published spatial transcriptomics on genetically perturbed and wild-type (WT) tumor lesions from a mouse model of human lung cancer (**Extended Data Fig. 2a**)[27]. We note the only available platform, as of now, for spatial CRISPR perturbations is via the 10x Genomics Visium platform which has a resolution of 1-10 cells per spot[28]. The ideology behind this *in vivo* validation was to 1) train on WT lesion, 2) simulate genetic perturbations in WT tissue, 3) compare model predicted transcriptomic differences with experimentally observed differences. To achieve this, we first isolated WT annotated lesions from the Visium slide and removed any spots proximal to experimentally perturbed spots, this mitigates information leakage issues (**Extended Data Fig. 2b, see Methods**). We used Celcomen to identify lung cancer specific gene-gene interaction modules from the WT lesions and then leveraged our model's generative component to predict spot-resolution transcriptomic profiles upon *in silico* Tgfbr2 knockout (KO). Consistent with Celcomen's accuracy in *ex vivo* human tissue, we once again observed strong agreement, significant positive correlation, between model predicted and experimentally observed transcriptomic changes, comparing Tgfbr2 KO and WT spots (**Extended Data Fig. 2c**). Through random permutation experiments, we confirmed that these correlations were unlikely to occur by chance which strongly supports Celcomen's ability to model spatial perturbations in a manner that agrees with experimentally derived ground truth (**Extended Data Fig. 2d**). In further support, we repeated this validation by comparing *in vivo* Jak2 KO and Celcomen *in silico* Jak2 KO spots with WT and, once again, observed significant positive correlation between Celcomen predicted and experimentally observed transcriptomic changes (**Extended Data Fig. 2e-f**). Thus, not only are we able to validate Celcomen's ability to recapitulate known biology in human tissue, but we are also able to *in vivo* validate Celcomen's spatial perturbation modeling capabilities in clinically relevant models of human disease.

## Discussion
The advent of single-cell resolution spatial transcriptomics has brought about a new paradigm to human cell mapping, allowing for spatial tissue atlases with unprecedented resolution[29–33]. While many computational methods have begun to address the phenotypic characterization of spatial transcriptomics, there remains a marked lack of works that take the next step forward and perform tissue level perturbation modeling[10,34]. There is a critical need for these methods in order to understand the mechanisms behind tissue dysfunction during disease states. Current works that broach this need are often uninterpretable, with putatively causal mechanisms hidden with a black box, or they are not mathematically robust leading to high variance in model outputs that limit their use.

Here, we present Celcomen, which addresses the need for perturbation modeling of tissue states with spatial context, while also providing highly interpretable results, through its disentanglement of cell intrinsic versus extrinsic gene regulation programs, and mathematical robustness through an identifiability guarantee. We confirm Celcomen's ability to disentangle and recover ground truth gene-gene interactions in real and self-simulated spatial transcriptomics data. These multi-faceted advances of Celcomen are likely to provide actionable insights into how human diseases cause tissue failure and allow for new testable hypotheses into the ways in which therapies provide patients with real clinical benefit. We anticipate that as technology continues to advance, the value of Celcomen and its future iterations will only continue to grow as it becomes more feasible to model disease state and more important to understand how.

## Methods

Spatial transcriptomics dataset curation and preprocessing

The fetal spleen datasets were curated from https://developmental.cellatlas.io/fetal-immune in log-normalized form, which explicitly indicates log-transformation and library size normalization[20]. The glioblastoma dataset was curated from 10x genomics at https://www.10xgenomics.com/datasets/ffpe-human-brain-cancer-data-with-human-immuno-oncology-profiling-panel-and-custom-add-on-1-standard and subjected to the same library size normalization, counts per million (CPM), and log-transformation, with a base of e; additionally, only genes that were expressed in at least 100 cells were kept. Due to the large size of the Xenium slide, a random square portion of the slide was chosen for analysis, this section is defined as cell centroid x-component > 6500 and < 7000 and cell centroid y-component > 8000 and < 8500. The entire fetal spleen slide was kept for each fetal slide sample as they are comparatively smaller than the original Xenium slide and post down-sampling, approximately the same size as the analyzed Xenium section. All data normalization were done using Scanpy (v1.9.8) in Python (v3.9.18)[35].

Simulations testing Celcomen's identifiability guarantees

Simulations were done in Python and completed by first generating a ground truth gene-gene interaction matrix. This was achieved by creating a n-genes by n-genes matrix of random values; for these experiments four genes were used. We then utilized Celcomen's generative module, Simcomen, to learn a spatially-resolved counts matrix reflective of the ground truth gene-gene interaction matrix. Comparisons to the randomly initialized count matrix are termed "Raw input" and those to the learned count matrix are termed "SCC output". To interrogate for self-consistency, we initialized Celcomen's inference module with a random gene-gene interaction matrix and asked it to utilize the learned count matrix from Simcomen to decipher the ground truth gene-gene interaction matrix. Comparisons to the Celcomen outputted gene-gene interaction matrix are termed "CCC output". Spearman correlation was used to compare the ground-truth gene-gene interaction values and the simulated-then-inferred gene-gene interaction values to test for model robustness and identifiability. For all exact parameter values utilized during the experiments, see the "analysis.simulations.ipynb" notebook in the reproducibility GitHub.

Biological testing of Celcomen's identifiability guarantees

Biological confirmation of Celcomen's identifiability guarantee was done by training two Celcomen inference module instances at the same time and comparing their derived gene-gene interaction results. The first model instance, which we call sample-specific, was trained only on one sample. The second model instance, which we call rest, was trained on the remaining samples. Thus, these two model instances are never trained on the same samples. Each model is trained to completion utilizing the same model hyperparameters, and their gene-gene interaction matrices are retrieved after the final epoch. We correlate a flattened version of their gene-gene interaction matrices using Spearman's correlation due to the possible non-linear nature of the matrices' values. We repeat this experiment for each of the samples in the fetal spleen dataset. The results across each sample's experiments are aggregated together and compared in a bar plot. We derived a "random" control to compare to by shuffling the order of the flattened gene-gene interaction matrices and computing a correlation of the shuffled values. Mann-Whitney U test is used to derive p-values and all p-values are labeled on plot. For the full code utilized, see the "analysis.biological.ipynb" notebook in the reproducibility GitHub.

Interferon knockout experiment on Xenium of human glioblastoma
Processed Xenium data was subjected to the inference module of Celcomen, CCE, and then these gene-gene interaction values were annotated as containing cytoplasmic, surface membrane (plasma membrane GO ID via GO cellular component), or secreted (extracellular space GO ID also via GO cellular component) genes according to their GO IDs from QuickGO[36]. *IFITM3* was knocked out in a randomly selected previously *IFITM3* positive cell. First neighbors were defined as less than 15 μm away and second neighbors were defined as less than 30 μm away. Changes in each gene's expression in each cell were calculated and these changes in expression pre- and post- perturbation were compared between different specified cellular subsets. These are the differential genes later used for differential expression analysis and pathway enrichment. Gene set enrichment analysis (GSEA) in R (v4.1.2) was utilized to perform pathway enrichment analysis on differentially post-perturbation affected genes. The interferon signature was derived directly from tissue by computing the differentially expressed genes between interferon high and low cells and taking the top 25, excluding the perturbed *IFITM3* as that would bias analyses. For the full model parameters and code utilized, see the "analysis.perturbation.ipynb" notebook in the reproducibility GitHub.

Counterfactual prediction validation via *in vivo* perturbed lung tumors
Spatial perturbation data was acquired from previously published Perturb-map technology, GSE193460[27]. Their processed spaceranger output and annotations were read in and wild-type (WT) lesions, as previously annotated, were identified and any spots that were within two degrees of a perturbation specific cluster were trimmed away; this was done via a <100 filter in spatial distance with the value of 100 visually acquired from a histogram of spot-spot spatial distances (i.e. distance of 100 was the second non-zero peak). Lesions were then fed into the Celcomen model to identify gene-gene relationships and the trained gene-gene interaction matrix was used by Simcomen for counterfactual predictions. In detail, each lesion was examined for Tgfbr2$^+$ spots and had a random positive spot knocked out (KO) in terms of Tgfbr2 expression. Simcomen then utilized the learned gene-gene interaction matrix to predict the whole transcriptome of every spot post perturbation. We then compared the change in expression in the KO spot compared to WT spots. Spearman correlation was used to compare model Tgfbr2 KO versus WT gene rankings with those directly derived from experimental Tgfbr2 KO spots and WT, i.e. the published data includes an *in vivo bona fide* Tgfbr2 KO lesion and this was used as ground truth. We derived "random" controls for each lesion by computing correlations on shuffled gene rankings of the observed and predicted differentials between Tgfbr2 KO and WT. Mann-Whitney U test is used to derive p-value when comparing observed lesion derived gene rankings with those from random shufflings. For the full code utilized, see the "analysis.biological.ipynb" notebook in the reproducibility GitHub.

## Data Availability
Data analyzed in this manuscript is previously published and is available from https://developmental.cellatlas.io/fetal-immune for human fetal spleen Visium, from https://www.10xgenomics.com/datasets/ffpe-human-brain-cancer-data-with-human-immuno-oncology-profiling-panel-and-custom-add-on-1-standard for human brain cancer Xenium, from https://www.ncbi.nlm.nih.gov/geo/query/acc.cgi?acc=GSE193460 for mouse perturbed lung tumor Visium[20,27].

## Code Availability
Celcomen is available as a python package under the GPL-3.0 license at https://github.com/stathismegas/celcomen. The code required for reproducing the analyses in this paper are at https://github.com/stathismegas/celcomen_reproducibility.

## Acknowledgments
We would like to thank Dr. Erick Armingol, Dr. Dinithi Sumanaweera, and Maciej Wiatrak for helpful feedback and discussions. This work was made possible in part by the Wellcome Trust (220540/Z/20/A). The funders had no role in study design, data collection and analysis, decision to publish, or preparation of the manuscript. Figure cartoons were created with BioRender.com.

## Author Contributions
S.M. conceived the theory of the model, implemented it in code and as a GitHub package, and wrote and edited the manuscript. D.G.C. performed the benchmarking on simulations and real data, the perturbation modeling and biological interpretation, and wrote and edited the manuscript. K.P. helped create conda environments for the project. M.E. gave feedback on the project. S.A.T. conceived the idea of disentangled representations in spatial data, provided key biological interpretation, and supervised the project.

## Competing Interests
In the past three years, S.A.T. has received remuneration for scientific advisory board membership from Sanofi, GlaxoSmithKline, Foresite Labs and Qiagen. S.A.T. is a co-founder and holds equity in Transition Bio and Ensocell. From January 8$^{th}$ of 2024, S.A.T. is a part-time employee of GlaxoSmithKline. The remaining authors declare no competing interests.

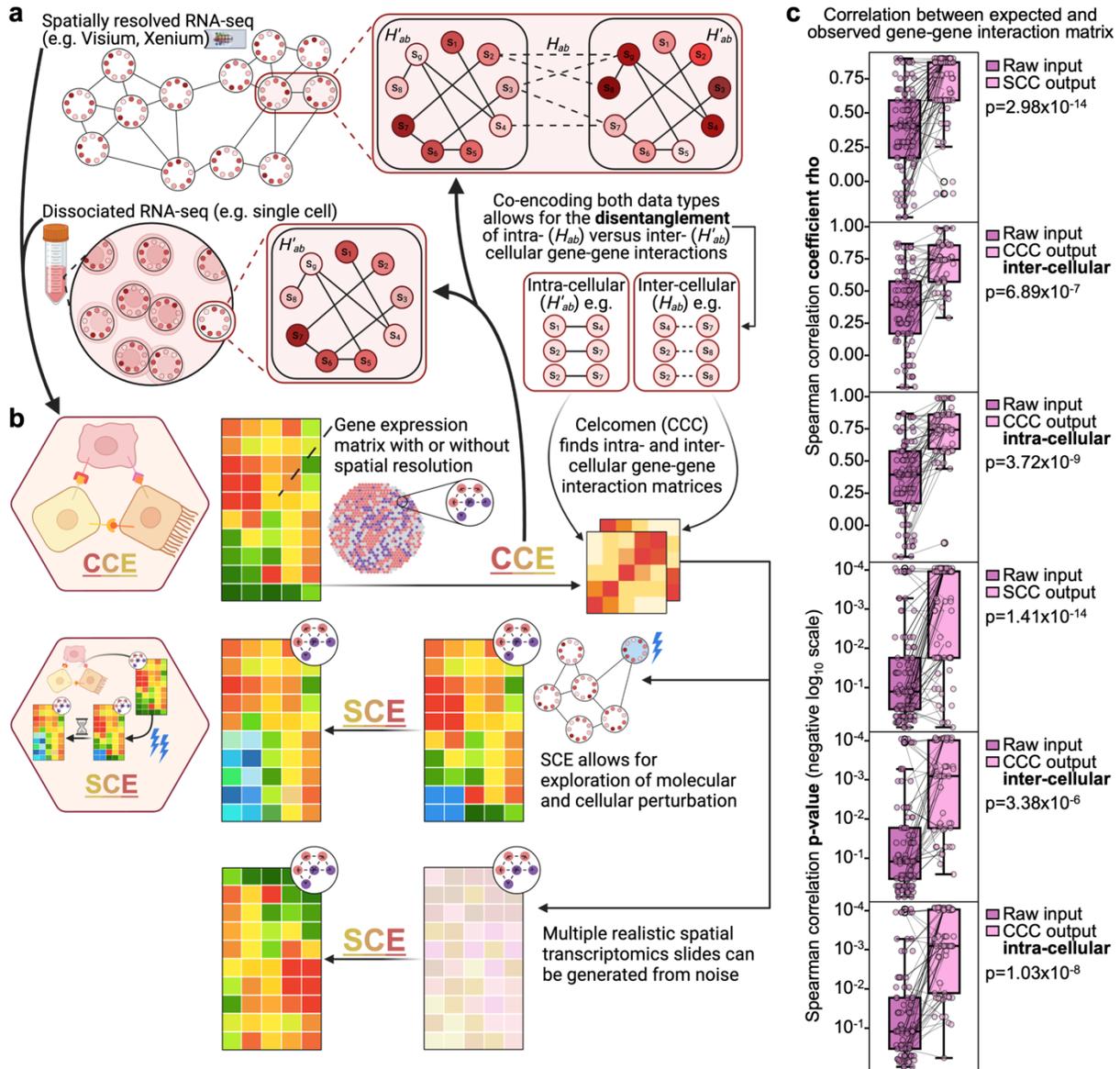

**Figure 1: Celcomen reproduces its identifiability guarantees in simulations**

a) Celcomen (CCC) can learn gene-gene relationships from either spatially resolved or dissociated RNA-seq data. The highlighted cell-cell pair, in spatial data, and individual cell, in scRNA-seq data, emphasizes how CCC can distinguish gene-gene interactions that are intra- ($H'_{ab}$) vs. inter- ($H_{ab}$) cellular gene-gene int.

b) Simcomen (SCC) leverages learned gene-gene relationships from CCC to model tissue behavior after cellular or genetic perturbation. SCC also possesses generative properties through its ability to create tissue-condition representative spatial transcriptomics data given an established matrix of gene-gene relationships.

c) Box plots with x-axis as the comparison, in detail, in magenta we compare the random count matrix with the ground truth and in light pink we compare the learned count matrix (SCC output) or gene-gene interaction matrix (CCC output) with ground truth. Y-axis is Spearman correlation coefficient rho (upper three) or p-value

(lower three). Mann-Whitney U-test p-values are labeled on the center right of each plot and the legend on the upper right of each plot labels each box's dataset.

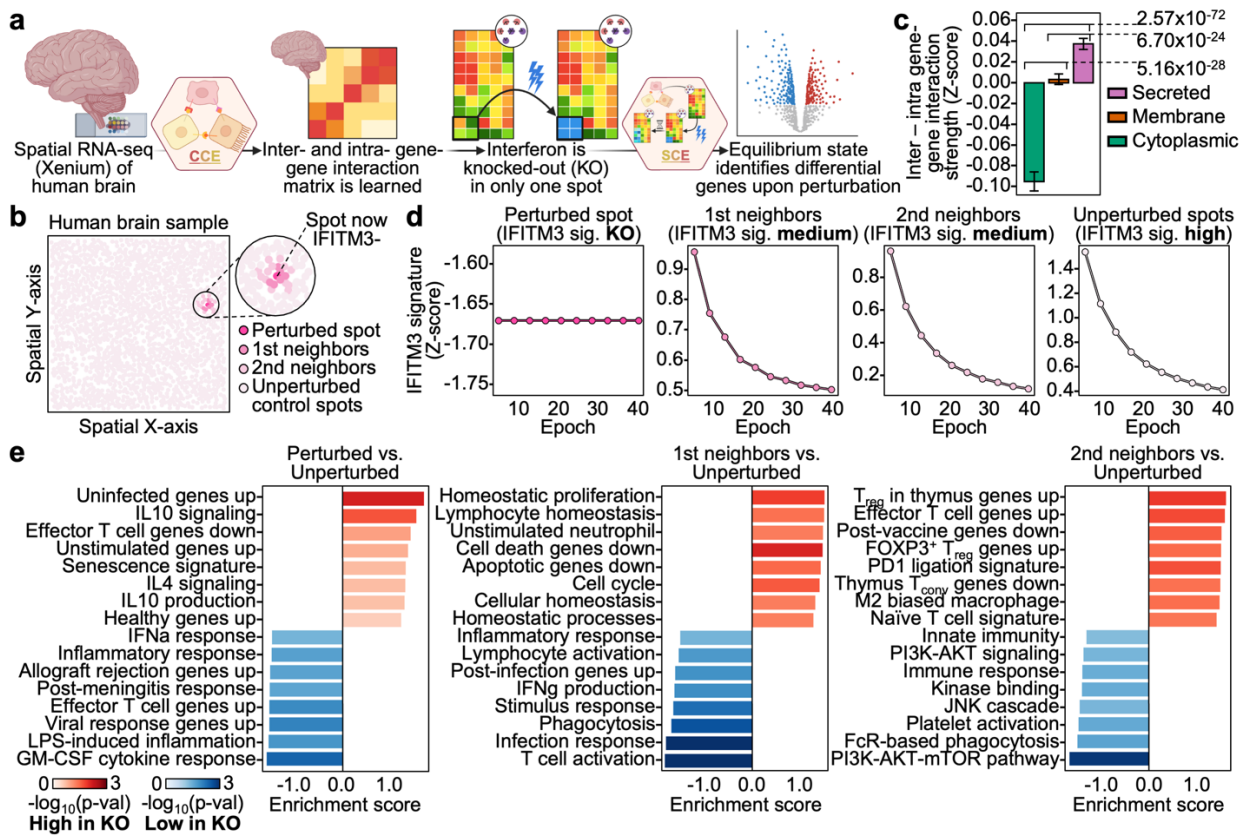

**Figure 2: Celcomen recapitulates known interferon knockout biology in human glioblastoma and disentangles intra- and inter- cellular gene-gene interactions**

a) Public spatially resolved RNA-seq data, Xenium, of the human brain during glioblastoma is inputted into CCC to derive intra- and inter- cellular gene-gene relationships. Interferon (IFN) signaling is *in silico* knocked out (KO) in a previously IFN[+] cell and SCC learns the local and global effects of this perturbation.

b) Cells within the region of interest are plotted with their spatial x- and y- coordinates. Legend on the lower right labels the identity of the perturbed cell in dark pink and its most proximal (1st) and lesser proximal (2nd) neighbors in lighter pink shades.

c) Bar plot with the x-axis as the subcellular localization of the gene as acquired from its gene ontology and the y-axis is the difference between the gene's inter- and intra- cellular gene-gene interaction terms. Mann-Whitney U-test p-values are labeled on the plot and error bars denote 95% confidence intervals.

d) Scatter plots with the x-axis as the epoch number and the y-axis as the interferon signature score of the given spot(s) at the specified epoch. Identity of the spot(s) of interest are labeled at the top of each plot and spot colors match those in (b).

e) Bar plots with the color as the pathway enrichment significance, see legend on the lower left, the x-axis as the enrichment score, and the y-axis for pathway names. Pathways were derived by first calculating pre- and post- perturbation changes in gene expression in each cell, then identifying differentially changed genes between spot(s) of interest and unperturbed controls, this provides a ranking of genes that were differentially upregulated or downregulated in the interferon KO cell, or its neighbors, as compared to the unperturbed control cells.

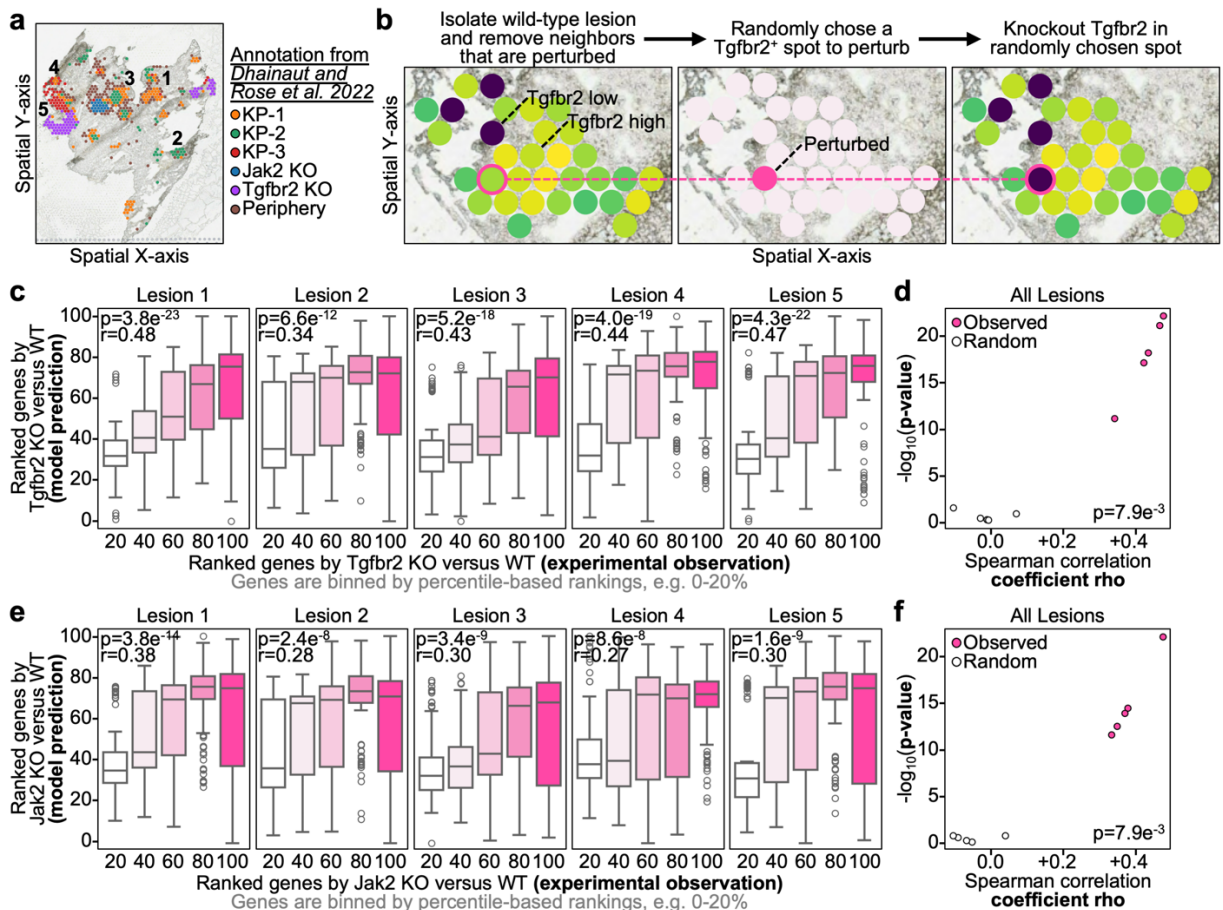

**Figure 3: Celcomen counterfactual predictions are validated *in vivo* in a clinically relevant lung cancer model**

a) Scatter plot on spatial axes with dots representing tumor lesions, color represents tumor cell phenotype, perturbation specific clusters are labeled with "KO" for knockout and wild-type tumors are labeled with "KP", see legend on right. Lesions of interest, large enough for modeling, are labeled with numbers.

b) Example workflow with a wild type (WT) lesion trimmed for spots within two-degrees of perturbed clusters, random Tgfbr2$^+$ spot has Tgfbr2 knocked out, our model then predicts the whole transcriptome that accompanies this perturbation.

c) Box plots, per lesion, with x-axis as the observed ranked differentially expressed genes (DEGs) between Tgfbr2 KO and WT and the y-axis as the model predicted gene ranking between our perturbed Tgfbr2 KO spot and wild type spots. Spearman correlation coefficient rhos and p-values are annotated on the plot.

d) Scatter plot with each dot representing a given tumor lesion with Tgfbr2 KO and the x-axis as the Spearman correlation coefficient rho and y-axis as the p-value, the color indicates if the correlation was computed on the lesion's observed gene rankings or a random shuffling of the gene rankings. Mann-Whitney U test p-value between observed and randomly shuffled correlations are annotated on the plot.

e) Box plots, per lesion, with x-axis as the observed ranked differentially expressed genes (DEGs) between Jak2 KO and WT and the y-axis as the model predicted gene ranking between our perturbed Jak2 KO spot and wild type spots. Spearman correlation coefficient rhos and p-values are annotated on the plot.

f) Scatter plot with each dot representing a given tumor lesion with Jak2 KO and the x-axis as the Spearman correlation coefficient rho and y-axis as the p-value, the color indicates if the correlation was computed on the lesion's observed gene rankings or a random shuffling of the gene rankings. Mann-Whitney U test p-value between observed and randomly shuffled correlations are annotated on the plot.

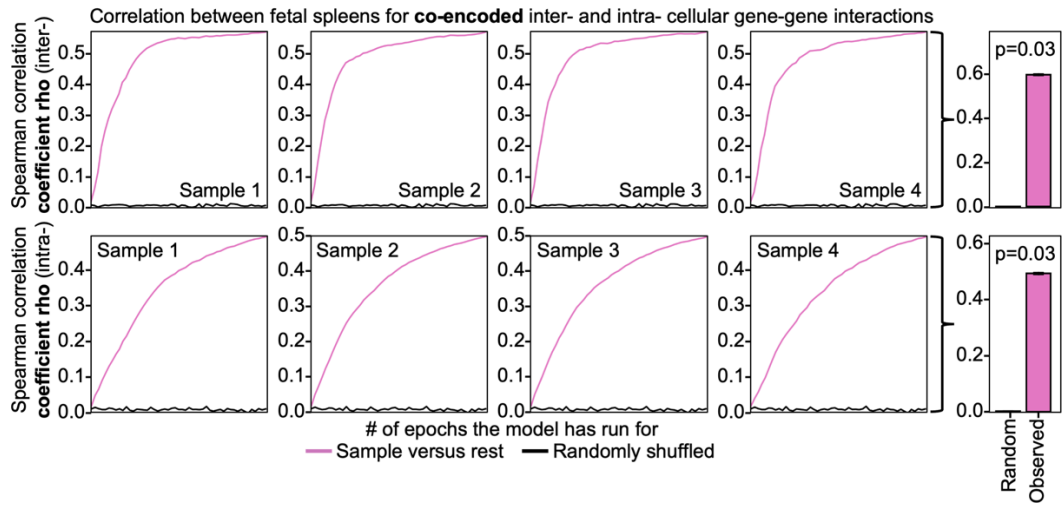

**Extended Data Figure 1: Celcomen recapitulates its identifiability guarantees through strong sample-to-sample correlation on real human samples**
Left: Line plots with the x-axis as epochs and y-axis as the Spearman correlation coefficients between the gene-gene interaction matrices of the model trained on the specified sample and the model trained on all other samples. The sample utilized for the sample specific model is annotated directly on the plot. The color of the line, see lower legend, indicates whether it represents comparisons between the two observed models, pink, or between a random shuffling of the two gene-gene interactions, black, to represent a null model. Right: Bar plots with the left black bar representing the average final Spearman correlation coefficient between randomly shuffled gene-gene interaction matrices of the sample specific model and model trained on all other samples, and the right pink bar representing the observed correlation. P-values are derived from Mann-Whitney U test and are annotated directly on plot. Error bars indicate standard error.

# Supplemental Notes: Causal disentanglement for spatial perturbation modeling

## 1 Notation

- $s_i^\alpha$, count values for spot/cell $i$ and gene $\alpha$,
- $\mathcal{H}$, Hamiltonian of a system,
- $Z = \sum_{\{s_i^\alpha\}} e^{\mathcal{H}(\{s_i^\alpha\})}$, the partition function,
- $\sum_{<i,j>_{nn}}$, sum over pairs of nodes $\{i,j\}$ that are nearest neighbors,
- $g_{\alpha\beta}$, Lagrange multiplier enforcing gene-gene correlations,
- q, the number of nearest neighbors (that we assume are interacting),
- $\mathcal{S}$, the entropy functional,
- S, the number of spots/nodes in the spatial graph,
- N, the number of features/genes in the graph,
- $J_{ij}$, the spatial adjacency matrix between spots/nodes in the graph,
- $\langle \rangle_P$, the average with respect to the probability distribution $P$,
- $\langle \rangle_{\text{exp}}$, the empirical/experimental average with respect to the observed samples,
- $P(s_i^\alpha) \in L^1$, the probability density of the count matrix of a spatial, transcriptomics experiment equals the matrix $s_i^\alpha$.

## 2 Motivation and Inspiration

Causal inference in machine learning aims to extract causal structures from observational data. As such, it stands in between correlation-based methods, and mechanistic models. Suppose for instance that biology imposes the co-localization of genes 1 and 2, and genes 2 and 3 in nearest neighbors. Since half of the time the nearest neighbor of a nearest neighbor is also a nearest neighbor, there will be (spurious) co-localization also of genes 1 and 3 in nearest neighbors. A causal model should be able to de-confound such spurious connections within spatial correlations, even without mechanistic data such as epigenetic information.

Inspiration for our work comes from the notion of force in physics. In broad strokes, we aim to learn the "least" number of forces (i.e. causal mechanisms which force the co-localization of pairs of genes) that can explain the observed spatial correlations of pairs of genes. "Least" here is meant in the sense of smallest entropy, not absolute number of forces (although we could additionally impose a L1 norm penalty on the force matrix).

In the Lagrangian formulation of classical physics, we think of time evolution of physical objects as an optimization problem (optimizing the action) such that certain constraints imposed by Lagrange multipliers are obeyed. One can show that Lagrange multipliers are equal to the force required to impose the corresponding constraint, which means that they are meaningful, physical quantities. At the same time, imposing the constrain via the Lagrange multipliers allows us to remain agnostic about the nature of the force (be it electromagnetism, gravity, or nuclear forces) that imposes the constraint. For an ant forced to walk on the surface of a table, this force (not letting it go through the table) happens to be electromagnetism, but we don't need to know this in advance to calculate its value.

Similarly in single cell genomics, measurements are valued in a high-dimensional gene expression space, but they often are hypothesised to lie on a much lower dimensional surface (see manifold hypothesis) due to biological mechanisms (already discovered or not) that "force" our measurements to lie on it. Uncovering such causal links is the first step to identifying the underlying molecular mechanisms. We use Lagrange multipliers to impose the observed co-localization of genes. Since, Lagrange multipliers are meaningful and physical quantities, finding them is likely to be a well-posed problem, leading to causally identifiable models. Moreover, if we could make the weights of the trained network equal to the Lagrange multipliers of our problem, then recovering the network weights would be easier.

## 3   Model Assumptions

Our model is the *unique* model that follows from three assumptions:

- that our model's expected gene-gene correlations across nearest neighbors match exactly the observed ones,
- that our model's expected gene-gene correlations within spots/cells match exactly the observed ones,

- that any other variables influencing the gene expression can be sufficiently modelled as white noise.

These three assumptions can be summarized in the following equation for the entropy

$$\mathcal{S}(P(\{s_i^\alpha\}), g_{\alpha\beta}, g'_{\alpha\beta}) = -\sum_{\{s_i^\alpha\}} P(\{s_i^\alpha\}) \log(P(\{s_i^\alpha\})) \\ + \sum_{\alpha,\beta} g_{\alpha\beta} (\langle \sum_{i,j\ nn} s_i^\alpha s_j^\beta \rangle_P - \langle \sum_{i,j\ nn} s_i^\alpha s_j^\beta \rangle_{\exp}) \\ + \sum_{\alpha,\beta} g'_{\alpha\beta} (\langle \sum_{i} s_i^\alpha s_i^\beta \rangle_P - \langle \sum_{i} s_i^\alpha s_i^\beta \rangle_{\exp}) \quad (1)$$

where $s_\alpha^i$ is the spatial gene expression and $P(\{s_\alpha^i\})$ is the probability distribution over possible spatial transcriptomics samples, and $g'_{\alpha\beta}$, $g_{\alpha\beta}$ are Lagrange multipliers that enforce our assumptions 1 and 2.

Our task now is to maximize the entropy functional 1 over all possible functions $P \in L^1(\mathbb{R}^{N \times S})$ and matrices $g_{\alpha\beta}$ and $g'_{\alpha\beta}$,

$$\max_{P,g,g'} \mathcal{S}(P(\{s_i^\alpha\}), g_{\alpha\beta}, g'_{\alpha\beta}). \quad (2)$$

## 4 Model Derivation

We should note that the optimization problem above is a particularly hard non-parametric problem, since it requires optimizing over not some numerical parameters but over the space of normalised functions. Relatedly, the entropy is not a function over numbers, but a functional over functions. Using functional calculus, we perform the maximization of the entropy functional in eq. 1 over all functions $P \in L^1(\mathbb{R}^{N \times S})$, to arrive at a simpler optimization problem over $g$ alone. This simpler optimization problem is more amenable by neural networks and will reveal the architecture our network should assume.

**Proposition 1** (Extremization over $P$). *The following two optimization problems are equivalent*

- *Maximizing the entropy functional in eq 1 over all possible functions $P \in L^1(\mathbb{R}^{N \times S})$ and matrices $g_{\alpha\beta}$ and $g'_{\alpha\beta}$*

$$\max_{P,g,g'} \mathcal{S}(P(\{s_i^\alpha\}), g_{\alpha\beta}, g'_{\alpha\beta}) \quad (3)$$

  *where $\mathcal{S}$ is given by 1,*

- *Minimizing the experimental/empirical log likelihood over matrices $g_{\alpha\beta}$ and $g'_{\alpha\beta}$*

$$\min_{g,g'} \langle \log P \rangle_{exp} = \min_{g,g'} \left( -\log Z(g_{\alpha\beta}) + g_{\alpha\beta} C_{\alpha\beta}^{exp} + g'_{\alpha\beta} C'^{exp}_{\alpha\beta} \right) \quad (4)$$

where $C_{\alpha\beta} = \sum_{i,j} s_{j\alpha} J_{ji} s_{i\beta}$ and $C'_{\alpha\beta} = \sum_i s_{i\alpha} s_{i\beta}$.

*Proof.* Optimizing a functional requires taking derivatives with respect to functions. In particular, using $\frac{\delta \int f(x)dx}{\delta f(y)} = \delta(x-y)$, we can maximize $S$ with respect to $P$:

$$0 = \frac{\delta \mathcal{S}}{\delta P(s')} = -\log P(\{s'^i_\alpha\}) - 1 + \sum_{\alpha,\beta} g_{\alpha\beta} \sum_{i,j\ nn} s'^\alpha_i s'^\beta_j + \sum_{\alpha,\beta} g'_{\alpha\beta} \sum_i s'^\alpha_i s'^\beta_i \quad (5)$$

$$\Rightarrow P(\{s^\alpha_i\}|\{g'_{\alpha\beta}, g_{\alpha\beta}\}) = \frac{e^{\mathcal{H}(\{s^\alpha_i\})}}{Z} \quad (6)$$

where we normalized the probability function and denote

$$\mathcal{H} = \sum_{\alpha\beta} \sum_i s^\alpha_i g'_{\alpha\beta} s^\beta_i + \sum_{\alpha\beta} \sum_{<i,j>nn} s^\alpha_i g_{\alpha\beta} s^\beta_j \quad (7)$$

$$= \sum_{\alpha\beta} \sum_i s^\alpha_i g'_{\alpha\beta} s^\beta_i + \sum_{\alpha\beta} \sum_{i,j} s^\alpha_i J_{ij} g_{\alpha\beta} s^\beta_j, \quad (8)$$

$$Z = \sum_{s^\alpha_i} e^{\mathcal{H}(\{s^\alpha_i\})}. \quad (9)$$

Maximizing with respect to the Lagrange multipliers $g_{\alpha\beta}, g'_{\alpha\beta}$ gives:

$$0 = \langle \sum_{i,j\ nn} s^\alpha_i s^\beta_j \rangle_P - \langle \sum_{i,j\ nn} s^\alpha_i s^\beta_j \rangle_{\exp}, \quad (10)$$

$$0 = \langle \sum_i s^\alpha_i s^\beta_i \rangle_P - \langle \sum_i s^\alpha_i s^\beta_i \rangle_{\exp}. \quad (11)$$

Moreover, by substituting 6 into 1 we get

$$\mathcal{S}(P(\{s^\alpha_i\}), g_{\alpha\beta}, g'_{\alpha\beta}) = \log Z - g_{\alpha\beta} \langle \sum_{i,j\ nn} s^\alpha_i s^\beta_j \rangle_{\exp} - g'_{\alpha\beta} \langle \sum_i s^\alpha_i s^\beta_i \rangle_{\exp} \quad (12)$$

$$= -\langle \log P(s) \rangle_{\exp} \quad (13)$$

Therefore maximizing $\mathcal{S}$ is equivalent to minimizing

$$\langle \log P \rangle_{\exp} = -\log Z(g_{\alpha\beta}) + g_{\alpha\beta} C^{\exp}_{\alpha\beta} + g'_{\alpha\beta} C'^{\exp}_{\alpha\beta} \quad (14)$$

where $C_{\alpha\beta} = \sum_{i,j} s_{j\alpha} J_{ji} s_{i\beta} = s^T @ J @ s$ in scanpy and numpy convention, or equivalently $g_{\alpha\beta} C^{\exp}_{\alpha\beta} = Tr(s@g@s^T @ J) = Tr(J@s@g@s^T)$. We now recognise $J@s@g$ as the message passing equation for a Graph Convolutional Network (GCN) [2]. □

In summary, our non-parametric optimization over $P$ tells us that the desired model architecture is a k-hop GCN [3] with a new and simpler loss function. In k-hop GCN, we separate the neighbors of a node in different levels (first neighbors, second neighbors, third neighbors, etc) and perform convolutional message passing from each of those levels to the, for each node in the graph.

# 5 Mean Gene Approximation

Despite the simpler simpler loss function of our k-hop GCN, it is still intractable to compute, because calculating the partition function (and its derivatives) requires summing over a large number of possible spatial transcriptomics datasets.

Several famous algorithms in machine learning circumvent computing the partition function in different ways. For instance, a contrastive learning approach essentially takes the ratio of probabilities, thereby cancelling out the partition function; optimization approaches cast the avoid the computation of the partition by considering maximum a-posteriori estimator [1]; and, score-based diffusion [4] uses score-matching to learn a model of the gradient of the log of the probability density function, which again avoids computing the partition function completely.

In this paper, we introduce a novel approximation to the partition function, inspired from physics, which has not been used before in spatial transcriptomics to our knowledge. This is a new Mean Field Theory approximation

$$s_k^\alpha = \bar{s}_k^\alpha + \delta s_k^\alpha = m^\alpha + \delta s_k^\alpha \qquad (15)$$

where we assume that the gene expression does not fluctuate much around the mean.

Using this, we can rewrite the exponent as

$$s_i^\alpha g_{\alpha\beta} s_j^\beta = g_{\alpha\beta}(\bar{s}_i^\alpha + \delta s_i^\alpha)(\bar{s}_j^\beta + \delta s_j^\beta) \qquad (16)$$

$$\approx g_{\alpha\beta}(\bar{s}_i^\alpha \bar{s}_j^\beta + \bar{s}_j^\beta \delta s_i^\alpha + \bar{s}_i^\alpha \delta s_j^\beta) \qquad (17)$$

$$= g_{\alpha\beta}(m^\alpha m^\beta + m^\beta(s_i^\alpha - m^\alpha) + m^\alpha(s_j^\beta - m^\beta)) \qquad (18)$$

$$= g_{\alpha\beta}(-m^\alpha m^\beta + m^\beta s_i^\alpha + m^\alpha s_j^\beta) \, . \qquad (19)$$

where in the second line we used the MFT approximation to neglect terms of order higher than 2, and

$$s_i^\alpha g'_{\alpha\beta} s_i^\beta = g'_{\alpha\beta}(-m^\alpha m^\beta + m^\beta s_i^\alpha + m^\alpha s_i^\beta) \, . \qquad (20)$$

This implies that the inter-cellular term in the exponent can be rewritten as

$$\sum_{\langle i,j \rangle} \sum_{\alpha,\beta} g_{\alpha\beta}(-m^\alpha m^\beta + m^\beta s_i^\alpha + m^\alpha s_j^\beta) = \frac{q}{2} \sum_i \sum_{\alpha,\beta} g_{\alpha\beta}(-m^\alpha m^\beta + m^\beta s_i^\alpha + m^\alpha s_i^\beta) \qquad (21)$$

where $q$ is the number of nearest neighbors that we assume are interacting, and

therefore

$$\mathcal{H} = \frac{q}{2} \sum_i \sum_{\alpha,\beta} g_{\alpha\beta}(-m^\alpha m^\beta + m^\beta s_i^\alpha + m^\alpha s_i^\beta) \tag{22}$$

$$+ \sum_i \sum_{\alpha,\beta} g'_{\alpha\beta}(-m^\alpha m^\beta + m^\beta s_i^\alpha + m^\alpha s_i^\beta) \tag{23}$$

$$= \sum_i \sum_{\alpha,\beta} \left(g'_{\alpha\beta} + \frac{q}{2} g_{\alpha\beta}\right)(-m^\alpha m^\beta + m^\beta s_i^\alpha + m^\alpha s_i^\beta) \tag{24}$$

since $g_{\alpha,\beta}$ is symmetric.

**Proposition 2.** *The following sum can be simplified as follows*

$$\sum_{\{s_i^\alpha\}} exp\left[\sum_i \sum_{\alpha,\beta} (\frac{q}{2} g_{\alpha\beta})(m^\beta s_i^\alpha + m^\alpha s_i^\beta)\right] = V_{\mathbb{S}^{n-1}} \left(\frac{e^{qH/2} - e^{-qH/2}}{qH/2}\right)^S \tag{25}$$

*where $S$ is the number of spots, $H_{\alpha\beta} = g_{\alpha\beta} + g_{\beta\alpha}$, $H = \sqrt{\sum_\beta (\sum_\alpha H_{\alpha\beta} m^\alpha)^2}$.*

*Proof.*

$$Z = \sum_{\{s_i^\alpha\}} \exp\left[\frac{q}{2} \sum_i \sum_{\alpha,\beta} g_{\alpha\beta}(m^\beta s_i^\alpha + m^\alpha s_i^\beta)\right] \tag{26}$$

$$= \sum_{\{s_i^\alpha\}} \exp\left[\frac{q}{2} \sum_i \sum_{\alpha,\beta} (g_{\beta\alpha} m^\alpha s_i^\beta + g_{\alpha\beta} m^\alpha s_i^\beta)\right] \tag{27}$$

$$= \sum_{\{s_i^\alpha\}} \exp\left[\frac{q}{2} \sum_i \sum_{\alpha,\beta} (g_{\beta\alpha} + g_{\alpha\beta}) m^\alpha s_i^\beta\right] \tag{28}$$

$$= \sum_{\{s_i^\alpha\}} \exp\left[\frac{q}{2} \sum_i \sum_{\alpha,\beta} H_{\alpha\beta} m^\alpha s_i^\beta\right] \tag{29}$$

$$= \prod_i \left(\int_{s_i \in \mathbb{S}^n} ds_i\right) \exp\left[\frac{q}{2} \sum_i \sum_{\alpha,\beta} H_{\alpha\beta} m^\alpha s_i^\beta\right] \tag{30}$$

$$= \prod_i \left(\int_{s_i \in \mathbb{S}^n} \exp\left[\frac{q}{2} \sum_i H s_i^1\right] ds_i\right) \tag{31}$$

$$= V_{\mathbb{S}^{n-1}} \prod_i \left(\int_0^\pi \exp\left[\frac{q}{2} \sum_i H \cos\theta\right] \sin\theta d\theta\right) \tag{32}$$

$$= V_{\mathbb{S}^{n-1}} \prod_i \left(\int_{-1}^1 \exp\left[\frac{q}{2} \sum_i Hu\right] du\right) \tag{33}$$

$$= V_{\mathbb{S}^{n-1}} \left(\frac{e^{qH/2} - e^{-qH/2}}{qH/2}\right)^S \tag{34}$$

$$\tag{35}$$

where $S$ is the number of spots, $H_{\alpha\beta} = g_{\alpha\beta} + g_{\beta\alpha}$, $H = \sqrt{\sum_{\beta}(\sum_{\alpha} H_{\alpha\beta} m^{\alpha})^2}$, and without loss of generality we assumed that the vector $\sum_{\alpha} H_{\alpha\beta} m^{\alpha}$ lies only along the first dimension. □

Now applying this proposition to our formula for the partition function, where we need to replace $g_{\alpha\beta} \to g_{\alpha\beta} + \frac{2}{q} g'_{\alpha\beta}$, gives

$$\log Z = -S \sum_{\alpha,\beta} \left( g'_{\alpha\beta} + \frac{q}{2} g_{\alpha\beta} \right) m^{\alpha} m^{\beta}$$
$$+ \log V_{\mathsf{S}^{n-1}}$$
$$+ S \log \frac{e^{H'/2} - e^{-H'/2}}{H'/2} \tag{36}$$

where $S$ is the number of spots, $H'_{\alpha\beta} = q g_{\alpha\beta} + q g_{\beta\alpha} + 2 g'_{\alpha\beta} + 2 g'_{\beta\alpha}$, $H' = \sqrt{\sum_{\beta}(\sum_{\alpha} H'_{\alpha\beta} m^{\alpha})^2}$

Using our 13, 4, 36, we have a complete formula for calculating the partition function and the only optimization remaining is over the Lagrange multipliers.

$$0 = \frac{\delta P(\{s_i^{\alpha}\})}{\delta g_{\alpha\beta}} \tag{37}$$
$$0 = \frac{\delta P(\{s_i^{\alpha}\})}{\delta g'_{\alpha\beta}} \tag{38}$$

In other words we want to look for the forces that are causing the observed spatial gene expression. Since the Lagrange multipliers/forces are meaningful physical variables, they naturally equip our model with identifiability as we show in the next section.

## 6 Identifiability

An important question we want to address is the identifiability of our model, i.e. whether there is a unique setting of the forces that leads to the observed correlations in the data. If the identifiability property holds then our model will naturally be robust and causal in the sense that it can de-confound spurious correlations and recover almost the Markov equivalence class of the causality diagram, which is the best any method without interventional data can do.

In mathematical terminology, we want to determine whether there is some gauge symmetry that allows different sets of parameters to produce the same probability distribution.

**Theorem 1** (Identifiability). *The model defined by equation 6 is identifiable in*

*the sense that*

$$\forall \{s_i^\alpha\}: \; P(\{s_i^\alpha\}|\{g_{\alpha\beta}, g'_{\alpha\beta}\}) = P(\{s_i^\alpha\}|\{h_{\alpha\beta}, h'_{\alpha\beta}\}) \tag{39}$$

$$\Rightarrow g_{\alpha\beta} = h_{\alpha\beta} \, and \, g'_{\alpha\beta} = h'_{\alpha\beta} \tag{40}$$

*Proof.* Let's pick $i$ to be a cell/node that has at least one neighbor. If there is not such a cell then there wouldn't be a cell communication problem to model.

$$P(\{s_i^\alpha\}|\{g_{\alpha\beta}, g'_{\alpha\beta}\}) = P(\{s_i^\alpha\}|\{h_{\alpha\beta}, h'_{\alpha\beta}\}) \tag{41}$$

$$\Rightarrow \frac{dP(\{s_i^\alpha\}|\{g_{\alpha\beta}, g'_{\alpha\beta}\})}{ds_i^\alpha} = \frac{dP(\{s_i^\alpha\}|\{h_{\alpha\beta}, h'_{\alpha\beta}\})}{ds_i^\alpha} \tag{42}$$

Then we pick $j$ to be any of the neighbors of cell $i$,

$$\frac{d^2 P(\{s_i^\alpha\}|\{g_{\alpha\beta}, g'_{\alpha\beta}\})}{ds_j^\beta ds_i^\alpha} = \frac{dP(\{s_i^\alpha\}|\{h_{\alpha\beta}, h'_{\alpha\beta}\})}{ds_j^\beta ds_i^\alpha} \tag{43}$$

$$\Rightarrow \left.\frac{d^2 P(\{s_i^\alpha\}|\{g_{\alpha\beta}, g'_{\alpha\beta}\})}{ds_j^\beta ds_i^\alpha}\right|_{s_{nn\,j}^\beta=0, s_i^\alpha=0} = \left.\frac{dP(\{s_i^\alpha\}|\{h_{\alpha\beta}, h'_{\alpha\beta}\})}{ds_j^\beta ds_i^\alpha}\right|_{s_{nn\,j}^\beta=0, s_i^\alpha=0} \tag{44}$$

$$\Rightarrow g_{\alpha\beta} = h_{\alpha\beta} \tag{45}$$

Alternatively, taking the second derivative with respect to the same cell $i$,

$$\frac{d^2 P(\{s_i^\alpha\}|\{g_{\alpha\beta}, g'_{\alpha\beta}\})}{ds_i^\beta ds_i^\alpha} = \frac{dP(\{s_i^\alpha\}|\{h_{\alpha\beta}, h'_{\alpha\beta}\})}{ds_i^\beta ds_i^\alpha} \tag{46}$$

$$\Rightarrow \left.\frac{d^2 P(\{s_i^\alpha\}|\{g_{\alpha\beta}, g'_{\alpha\beta}\})}{ds_i^\beta ds_i^\alpha}\right|_{s_i^\beta=0, s_i^\alpha=0} = \left.\frac{dP(\{s_i^\alpha\}|\{h_{\alpha\beta}, h'_{\alpha\beta}\})}{ds_i^\beta ds_i^\alpha}\right|_{s_i^\beta=0, s_i^\alpha=0} \tag{47}$$

$$\Rightarrow g'_{\alpha\beta} = h'_{\alpha\beta} \tag{48}$$

□

# 7 Simcomen: Generation Module

Our model offers a mathematically robust way of learning the distribution of spatial transcriptomics samples such that there is a 1-1 correspondence between a configuration of forces and the learned distribution of spatial transcriptomics samples.

Generating new samples from the learned distribution is classic problem that can be addressed for instance by Markov Chain Monte Carlo Methods. However, given the high dimensionality of the space of spatial transcriptomics, MCMC can be very computationally expensive. Therefore in our generation module, called Simulated Communication Energy, we produce new samples in a denoising like approach by fixing the parameters of our model and optimizing the likelihood of the given data either sampled from a different distribution, e.g. having Gaussian represent noise, or to generate counterfactual samples, e.g. we intervene on a spot or cell and from that starting point we find the most likely spatial distribution of gene expression values under the learned distribution.

## 8 One-gene inter-cellular communication leads to a convex optimization problem

In this section we study the optimization problem of $Z$ when there is only one gene. The main results of this section is the following theorem.

**Proposition 3** (MFT approximation leads to convex optimization problem). *When the target space is of dimension one (i.e. there is only one feature), the minimization problem of $\min_{g_{\alpha\beta}} \langle \log P \rangle_{exp}$, where the partition function is approximated by MFT (see eq. 36), is a convex problem.*

*Proof.* For one gene we have

$$H = \sqrt{\sum_\beta (\sum_\alpha H_{\alpha\beta} m^\alpha)^2} = |H_{11}m| = 2|g_{11}m| \qquad (49)$$

and therefore

$$\log Z(g_{11}) = -\frac{q}{2}Sg_{11}m^2 + \log V_{\mathsf{S}^{n-1}} + S\log \frac{e^{q|g_{11}m|} - e^{-q|g_{11}m|}}{q|g_{11}m|} \qquad (50)$$

$$\Rightarrow \langle \log P \rangle_{\exp} = -\log Z(g_{11}) + g_{11} C_{11}^{\exp} \qquad (51)$$

$$= \frac{q}{2}Sg_{11}m^2 - \log V_{\mathsf{S}^{n-1}} - S\log \frac{e^{q|g_{11}m|} - e^{-q|g_{11}m|}}{q|g_{11}m|} + g_{11} C_{11}^{\exp} \qquad (52)$$

$$\Rightarrow \frac{d}{dg_{11}} \langle \log P \rangle_{\exp} = \frac{q}{2}Sm^2 - S\left(\frac{q|m|(\pm 1)}{\tanh(q|g_{11}m|)} - \frac{(\pm 1)}{|g_{11}|}\right) + C_{11}^{\exp} \qquad (53)$$

$$\Rightarrow \frac{d}{dg_{11}} \langle \log P \rangle_{\exp} = \frac{q}{2}Sm^2 - S\left(\frac{q|m|}{\tanh(q|m|g_{11})} - \frac{1}{g_{11}}\right) + C_{11}^{\exp} \qquad (54)$$

where the + sign is for $g_{11} > 0$ and the minus sign for $g_{11} < 0$.

As we can see in the figure below, $-Sqm \leq -S\left(\frac{q|m|}{\tanh(q|m|g_{11})} - \frac{1}{g_{11}}\right) \leq Sqm$ eq. 54 has one root only if $\frac{m}{2} < 1$, otherwise it is always positive. Another necessary constraint similarly exists for $C_{11}^{\exp}$ which tells us that the mean field theory approximation might sometimes need to be adjusted. Another direct prediction from equation 54 is that there is at most one root, and therefore our optimization problem is convex.

□

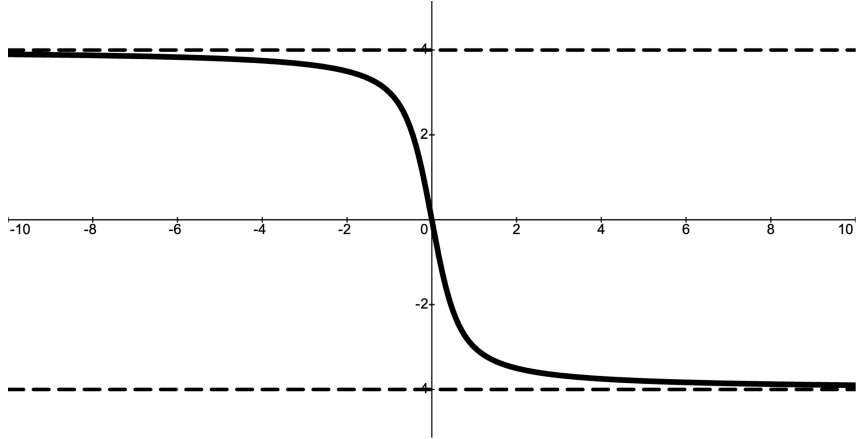

The second derivative is

$$\Rightarrow \frac{d^2}{d(g_{11})^2}\langle \log P \rangle_{\exp} = -S\bigg(-q^2|m|^2\frac{1}{\sinh^2(q|m|g_{11})} + \frac{1}{(g_{11})^2}\bigg) \qquad (55)$$